# MULTIMODAL BIOMETRIC SYSTEMS – STUDY TO IMPROVE ACCURACY AND PERFORMANCE

K.Sasidhar[1], Vijaya L Kakulapati[2], Kolikipogu Ramakrishna[3] & K.KailasaRao[4]

[1]Department of Master of Computer Applications, MLRCET, Hyderabad, India
[2]Department of Computer Science & Engineering, JNT University, Hyderabad, India
vldms@yahoo.com

[3]Department of Computer Science & Engineering, JNT University, Hyderabad, India
krkrishna.csit@gmail.com

[4]Department of Master of Computer Applications, MLRCET, Hyderabad, India

## ABSTRACT

*Biometrics is the science and technology of measuring and analyzing biological data of human body, extracting a feature set from the acquired data, and comparing this set against to the template set in the database. Experimental studies show that Unimodal biometric systems had many disadvantages regarding performance and accuracy. Multimodal biometric systems perform better than unimodal biometric systems and are popular even more complex also. We examine the accuracy and performance of multimodal biometric authentication systems using state of the art Commercial Off- The-Shelf (COTS) products. Here we discuss fingerprint and face biometric systems, decision and fusion techniques used in these systems. We also discuss their advantage over unimodal biometric systems.*

## 1. INTRODUCTION

The Multimodal biometric systems are providing identification and human security over last few decades. Due to this reason MBS are adapted to many fields of applications. Some of these multimodal systems are human computer dialog interaction based systems where the user interacts with the PC through voice or vision or any other pointing device in order to complete a specific task. Multimodal biometric systems are those which utilize, or are capability of utilizing, more than one physiological or behavioral characteristic for enrollment, verification, or identification. A biometric system is essentially a pattern recognition system. This system measure and analyzes human body Physiological characteristics, such as fingerprints, eye retinas and irises, voice patterns, facial patterns and hand measurements for authentication purposes or behavioural characteristics. The biometric identifiers cannot be misplaced. In spite of inherent advantages, unimodal biometric solutions also have limitations in terms of accuracy, enrolment rates, and susceptibility to spoofing. This limitation occurs in several application domains, example is face recognition. The accuracy of face recognition is affected by illumination and facial expressions. The biometric system cannot eliminate spoof attacks.

Example is finger print spoofing with rubber. A recent report by the National Institute of Standards and Technology (NIST) to US concluded that approximately two percent of the population does not have a legible fingerprint [1]. Inspite of using unimodal biometric system that have poor performance and accuracy, we study and propose a new approach to the multimodal biometric system. This new Multimodal biometric systems perform better than unimodal biometric systems and are popular even more complex also

## 2. MULTI MODAL BIOMETRIC SYSTEM

Multi modal biometric systems utilize more than one physiological or behavioural characteristic for enrolment, verification or identification. The NIST report recommends a system employing multiple biometrics in a layered approach. The reason to combine different modalities is to improve recognition rate. The aim of multi biometrics is to reduce one or more of the following:

- False accept rate (FAR)

- False reject rate (FRR)

- Failure to enroll rate (FTE)

- Susceptibility to artefacts or mimics

Multi modal biometric systems take input from single or multiple sensors measuring two or more different modalities of biometric characteristics. For example a system with fingerprint and face recognition would be considered "multimodal" even if the "OR" rule was being applied, allowing users to be verified using either of the modalities [4].

### 2.1. Multi algorithmic biometric systems

Multi algorithmic biometric systems take a single sample from a single sensor and process that sample with two or more different algorithms.

### 2.2. Multi-instance biometric systems

Multi-instance biometric systems use one sensor or possibly more sensors to capture samples of two or more different instances of the same biometric characteristics. Example is capturing images from multiple fingers.

### 2.3. Multi-sensorial biometric systems

Multi-sensorial biometric systems sample the same instance of a biometric trait with two or more distinctly different sensors. Processing of the multiple samples can be done with one algorithm or combination of algorithms. Example face recognition application could use both a visible light camera and an infrared camera coupled with specific frequency.

## 3. FUSION IN MULTIMODAL BIOMETRIC SYSTEMS

A Mechanism that can combine the classification results from each biometric channel is called as biometric fusion. We need to design this fusion.

Multimodal biometric fusion combines measurements from different biometric traits to enhance the strengths. Fusion at matching score, rank and decision level has been extensively studied in the literature. Various levels of fusion are: Sensor level, feature level, matching score level and decision level.

1. Sensor level Fusion:

We combine the biometric traits taken from different sensors to form a composite biometric trait and process.

2. Feature level Fusion:

Signal coming from different biometric channels are first pre-processed, and Feature vectors are extracted separately, using specific algorithm and we combine these vectors to form a composite feature vector. This is useful in classification.

3. Matching score level fusion:

Rather than combining the feature vector, we process them separately and individual matching score is found, then depending on the accuracy of each biometric matching score which will be used for classification.

4. Decision level fusion:

Each modality is first pre-classified independently.

Multimodal biometric system can implement any of these fusion strategies or combination of them to improve the performance of the system; different levels of fusion are shown in below figure-I

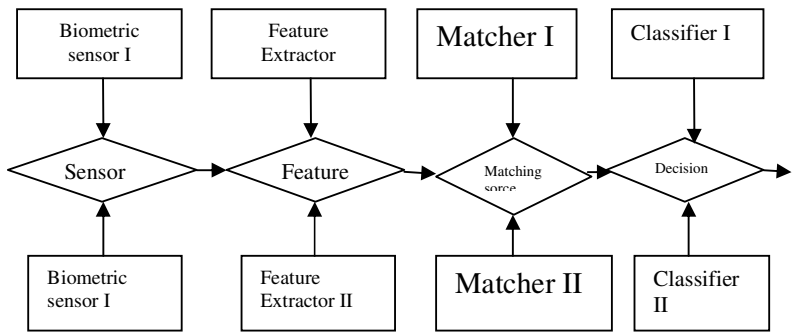

Figure –I. Fusion levels in multi modal biometric systems

## 3.1. Architecture

Here we discuss existing architecture. In literature Jain and Ross has discussed a multimodal biometric system using face and finger print and proposed various levels of combinations of the fusion. This is shown in Figure-II

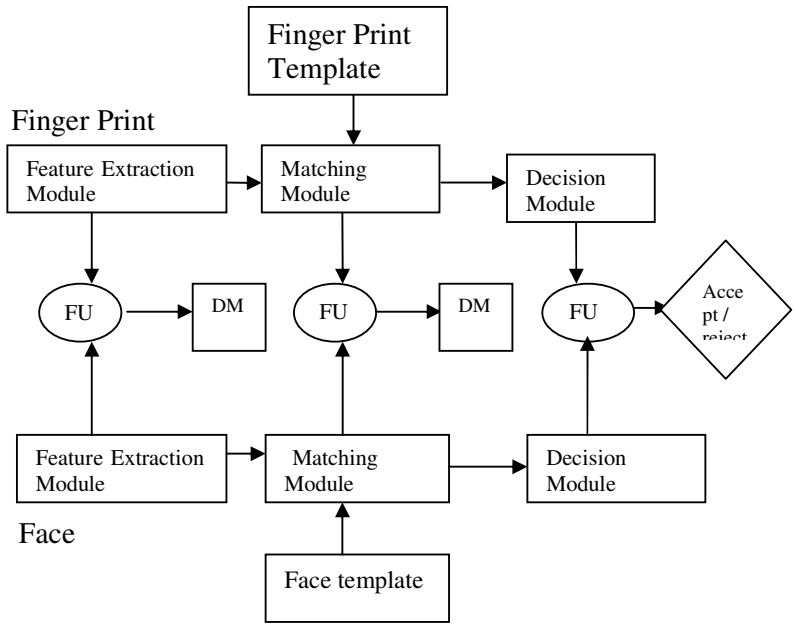

Figure – II Multimodal biometric system using face and fingerprint
(FU – fusion DM – Decision Module)

Yan and Zang have proposed a correlation Filter bank based fusion for multimodal biometric system; they used this approach for Face & Palm print biometrics. In Correlation Filter Bank, the unconstrained correlation filter trained for a specific modality is designed by optimizing the overall original correlation outputs. Therefore, the differences between Face & Palm print modalities have been taken into account and useful information in various modalities is fully exploited. PCA was used to reduce the dimensionality of feature set and then the designed correlation filter bank (CFB) was used for fusion. Fig. III shows the fusion network architecture proposed by them, the recognition rates achieved are in the range 0.9765 to 0.9964 with the proposed **method**

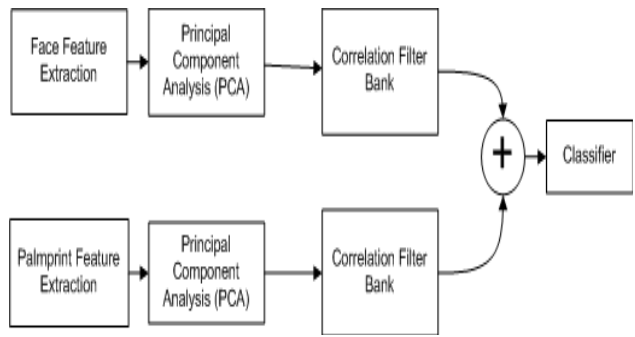

Figure III:  Correlation Filter bank based fusion

## 3.2. Normalization

In this section, we present well-known normalization methods. We denote a raw matching score as s, from the set S of all scores for that matcher, and the corresponding normalized score as s'.

**Min-Max**      :   *s' = (s - min) / (max-min)*

**Zscore**        :   *s' = (s - mean)/(standard deviation)*

**MAD**          :   *s' = (s - median)/constant (median | - median|)*

**tanh**          :   *s' = .5[ tanh ( .01(s - mean)/(standard deviation))+1]*

Normalization addresses the problem of incomparable classifier output scores in different combination classification systems.

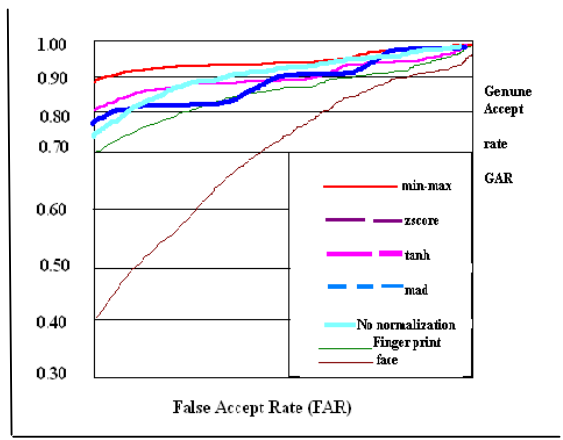

.                    Figure 4: simple sum rule with different normalizations

# 4. EXPERIMENTS

**ROC (Receiver Operating Characteristic) curve implementation**:

Performance statistics are computed from the real and fraud scores. Real scores are those that result from comparing elements in the target and query sets of the same subject. Fraud scores are those resulting from comparisons of different subjects. Use each fusion score as a threshold and compute the false-accept rate (FAR) and false-reject rate (FRR) by selecting those fraud scores and genuine scores, respectively, on the wrong side of this threshold and divide by the total number of scores used in the test. A mapping table of the threshold values and the corresponding error rates (FAR and FRR) are stored. The complement of the FRR (1 – FRR) is the Genuine accept-rate (GAR). The GAR and the FAR are plotted against each other to yield a ROC curve, a common system performance measure. We choose a desired operational point on the ROC curve and uses the FAR of that point to determine the corresponding threshold from the mapping table.  Figure 4 shows a ROC (Receiver Operating Characteristic) curve for the simple sum fusion rule with various normalization techniques. Clearly the use of these fusion and normalization techniques enhances the performance significantly over the single-modal face or fingerprint classifiers. For example, at a FAR of 0.1% the simple sum fusion with the min-max normalization has a GAR of 94.9%, which is considerably better than that of face, 75.3%, and fingerprint, 83.0%. Also, using any of the normalization techniques in lieu of not normalizing the data proves beneficial. The simplest normalization technique, the min-max, yields the best performance in this example.

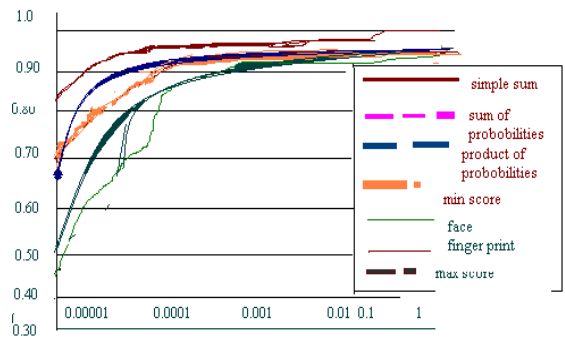

Figure 5  Min-Max Normalization with different fusions

Figure 5 illustrates the results of Min-Max normalization for a spectrum of fusion methods. The simple sum fusion method yields the best performance over the range of FARs. Interestingly, the Genuine-Accept Rate for sum and product probability rules falls off dramatically at a lower FAR. GAR for the spectrum of normalization and fusion techniques at FARs of 1% and 0.1% respectively. At 1% FAR, the sum of probabilities fusion works the best. However, these results do not hold true at a FAR of 0.1%. The simple sum rule generally performs well over the range of normalization techniques. These results demonstrate the utility of using multimodal biometric systems for achieving better matching performance. They also indicate that the method chosen for fusion has a significant impact on the resulting performance. In operational biometric systems, application requirements drive the selection of tolerable error rates and in both single-modal and multimodal biometric systems, implementers are forced to make a trade-off between usability and security. In operational biometric systems, application requirements drive the selection of tolerable error rates and in both single-modal and multimodal biometric systems, implementers are forced to make a trade-off between usability and security.

## 5. CONCLUSION

A framework was established with assessing the performance of multimodal biometric systems. We have examined relatively large face and fingerprint data sets over a spectrum of normalization and fusion techniques. The results of this study shows multimodal biometric systems better perform than uni-modal biometric systems. An additional advantage of fusion at this level is that existing and proprietary biometric systems do not need to be modified, allowing for a common middleware layer to handle the multimodal applications with a small amount of common information. Future scope is to investigate alternative normalization and fusion methods. single-mode biometrics testing, has concluded  that to accurately evaluate the performance of biometric systems, tests must be performed with data sets on the order of tens-of-thousands subjects and that no inferences be drawn from tests conducted on small subject populations to assess system scalability. Thus, future plans include expanding the test databases to attain these larger sizes. In addition, to assess the feasibility of such systems for large-scale deployments, we will perform these tests using COTS products.

## Authors

1.       K.Sasidhar, M.Tech , MISTE, He is pursuing Ph.D in Computer Science and Engineering. He has around 13 years of Professional Teaching Experience in Various Engineering colleges. Now he is working as HOD, Dept. of MCA, MLRIT , Hyderabad.For his credit he has Published one International Journal and Participated in 6 conferences and Seminars

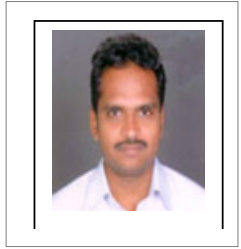

2.       Vijaya Lakshmi Kakulapati, M.Tech(IT) and Pursuing Ph.D in field of Data Mining from Jawaharlal Nehru Technological University Hyderabad, She has 18 years of teaching Experience in various Educational Institutions. For her credit she has more than 5 renewed publications in National and International Conference/Journals.

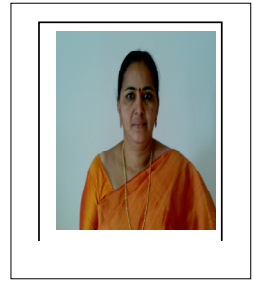

3.       Kolikipogu Ramakrishna, M.Tech (Software Engineering) ,LMISTE and Pursuing Ph.D in Computer Science & Engineering in the area of Information Retrieval from JNTU Hyderabad. He has 7 years of Professional teaching experience in various Engineering colleges. For his credit he has published 2 National symposium papers & Two International Conference/Journal Papers.

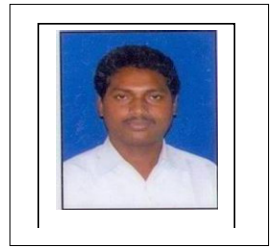

4.       K.Kailasa Rao, M.Tech (IT), MISTE and Pursuing his Ph.D in Computer Science and Engineering. He has 28 years of Experience including 19 years Industry and 9 years teaching. He is currently working as HOD, Department of CSE, Malla Reddy Institute of Technology, Hyderabad, India. He published one International Journal, 2 Research papers and 3 National Conference papers to his credit.

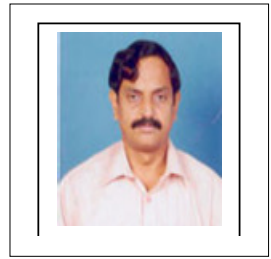